\documentclass[a4paper, conference]{IEEEtran}

\usepackage{cite}
\usepackage{amsmath,amssymb,amsfonts}
\usepackage{algorithmic}
\usepackage{graphicx}
\usepackage{textcomp}
\usepackage{xcolor}
\usepackage{paralist}
\usepackage{eurosym}
\usepackage{flushend}
\usepackage{url}
\usepackage[utf8]{inputenc}
\usepackage[colorlinks=true]{hyperref}
\usepackage{soul}
\usepackage{caption}
\usepackage[list=true]{subcaption}
\usepackage{booktabs}
\usepackage{tabu}
\usepackage{wrapfig}
\usepackage{arydshln}
\usepackage[normalem]{ulem}
\usepackage{multirow}
\usepackage{siunitx}
\usepackage{adjustbox}
\usepackage{array}
\usepackage{diagbox}

\begin{document}

\title{VAE-based Feature Disentanglement for Data Augmentation and Compression in Generalized GNSS Interference Classification}

\author{\IEEEauthorblockN{Lucas Heublein\IEEEauthorrefmark{1},
    Simon Kocher\IEEEauthorrefmark{1},
    Tobias Feigl\IEEEauthorrefmark{1},
    Alexander Rügamer\IEEEauthorrefmark{1},
    Christopher Mutschler\IEEEauthorrefmark{1},
    \underline{Felix Ott}\IEEEauthorrefmark{1}}
    \IEEEauthorblockA{\IEEEauthorrefmark{1}Fraunhofer Institute for Integrated Circuits IIS, 90411 Nürnberg, Germany}
    \IEEEauthorblockA{\{lucas.heublein, simon.kocher, tobias.feigl, alexander.ruegamer, christopher.mutschler, felix.ott\}@iis.fraunhofer.de}
}

\IEEEoverridecommandlockouts
\IEEEpubid{\makebox[\columnwidth]{
979-8-3315-1113-5/25/\$31.00~\copyright2025
IEEE \hfill} \hspace{\columnsep}\makebox[\columnwidth]{ }}

\maketitle

\begin{abstract}
Distributed learning and Edge AI necessitate efficient data processing, low-latency communication, decentralized model training, and stringent data privacy to facilitate real-time intelligence on edge devices while reducing dependency on centralized infrastructure and ensuring high model performance. In the context of global navigation satellite system (GNSS) applications, the primary objective is to accurately monitor and classify interferences that degrade system performance in distributed environments, thereby enhancing situational awareness. To achieve this, machine learning (ML) models can be deployed on low-resource devices, ensuring minimal communication latency and preserving data privacy. The key challenge is to compress ML models while maintaining high classification accuracy. In this paper, we propose variational autoencoders (VAEs) for disentanglement to extract essential latent features that enable accurate classification of interferences. We demonstrate that the disentanglement approach can be leveraged for both data compression and data augmentation by interpolating the lower-dimensional latent representations of signal power. To validate our approach, we evaluate three VAE variants -- vanilla, factorized, and conditional generative -- on four distinct datasets, including two collected in controlled indoor environments and two real-world highway datasets. Additionally, we conduct extensive hyperparameter searches to optimize performance. Our proposed VAE achieves a data compression rate ranging from 512 to 8,192 and achieves an accuracy up to 99.92\%.
\end{abstract}
\IEEEpeerreviewmaketitle

\section{Introduction}
\label{label_introduction}

Distributed learning involves training ML models across multiple devices or nodes, enabling data processing closer to its source and reducing reliance on centralized servers~\cite{threadgill_gerstlauer,shrivastava_isik_li}. Edge AI refers to deploying ML-based algorithms directly on edge devices, such as smartphones or IoT devices, allowing for real-time data processing and decision-making with minimal latency~\cite{meuser_loven}. Implementing ML models in these contexts necessitates lightweight algorithms for data preprocessing, efficient model architectures to accommodate limited computational resources, and robust mechanisms to ensure data privacy and security~\cite{gill_golec,wang_nepal_moore}. In distributed learning, compressing data is essential to reduce communication latency and bandwidth usage, thereby enhancing training efficiency~\cite{rebai_ojewale_ullah,liang_zhang_lu,raichur_ion_gnss}.

In GNSS interference monitoring, compact devices with efficient ML methods are crucial for deployment in diverse environments while ensuring minimal power consumption~\cite{rijnsdorp_zwol}. Reliable ML algorithms must process large datastreams in real time to detect and classify interference effectively~\cite{li_huang_lang}. The ability to handle vast amounts of data with minimal latency ensures timely responses to signal disruptions, improving navigation and positioning accuracy~\cite{merwe_franco}. Within the context of GNSS interference detection, a diverse range of ML techniques has been explored. These include approaches based on pseudo-labeling~\cite{heublein_feigl_posnav}, few-shot learning~\cite{ott_heublein_icl}, time- and frequency-domain sensor fusion~\cite{brieger_ion_gnss}, continual learning~\cite{raichur_heublein}, federated learning~\cite{gaikwad_heublein}, and transfer learning~\cite{swinney_woods}. Various ML models, such as convolutional neural networks (CNNs)~\cite{ferre_fuente,mehr_dovis}, support vector machines (SVMs)~\cite{li_huang_lang}, random forests~\cite{xu_ying_li}, and others~\cite{ghanbarzade_soleimani,mohanty_gao,iqbal_aman_sikdar,mascher_laller}, have been evaluated, while language models have also been leveraged for explainability~\cite{manjunath_heublein}. These techniques aim to mitigate domain shifts, enhance feature representations, and facilitate adaptation between simulated and outdoor scenarios. However, research on GNSS-based compression and feature disentanglement remains limited~\cite{ghobadi_spogli,reda_mekkawy}.

While compressing fine-grained data is essential, traditional compression techniques~\cite{libutti_igul} often degrade the resolution of interference signals, thereby reducing their informational value. A decline in signal representation quality compromises the performance of detection and classification systems, ultimately hindering the localization and mitigation of interference sources. Although advanced compression algorithms~\cite{gopinath_ravisankar} can retain more information, their increased computational complexity raises system costs and limits their feasibility for real-time deployment. Extracting relevant features from GNSS snapshots through feature disentanglement enables data compression specifically for training purposes, thereby minimizing the amount of transmitted information. This approach supports low-latency communication while maintaining or even improving accuracy in GNSS applications~\cite{rakhmanov_wiseman}. VAEs~\cite{kingma_welling,kim_mnih} and generative adversarial networks (GANs)~\cite{bao_chen_wen} are commonly used for disentanglement by learning structured latent representations that separate independent factors of variation in data.

\begin{figure}[!t]
    \centering
    \includegraphics[trim=0 6 0 6, clip, width=1.0\linewidth]{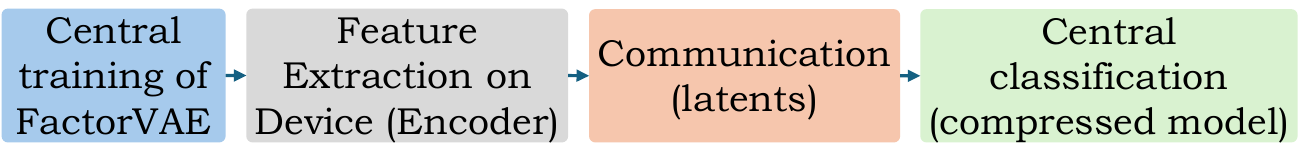}
    \caption{Pipeline for compressed classification.}
    \label{figure_pipeline}
\end{figure}

\textbf{Contributions.} The primary objective of this work is to compress GNSS snapshot data for a distributed system (see Figure~\ref{figure_pipeline}). We introduce a disentanglement approach to extract essential features from GNSS signals, including interference class, signal power, bandwidth, and distance to the interference source, by leveraging VAEs. To achieve this, we propose a method based on a VAE~\cite{kingma_welling}, a factorized VAE~\cite{kim_mnih}, and CVAE-GAN~\cite{bao_chen_wen} to learn an optimal latent variable representation. This compressed, lower-dimensional representation enables low-latency data transmission, efficient model training in a distributed system, and rapid inference. Additionally, we introduce a technique to interpolate key latent variables for data augmentation, thereby enhancing model generalization. Our proposed VAE achieves a high compression rate while maintaining high classification accuracy.

\textbf{Outlook.} Section~\ref{label_related_work} offers a review of existing literature on compression and data augmentation for GNSS data. Section~\ref{label_method} introduces the proposed vanilla, factorized, and conditional VAE. The datasets employed in this study are described in Section~\ref{label_experiments}, while Section~\ref{label_evaluation} presents a summary of the evaluation results. Finally, Section~\ref{label_conclusion} concludes the paper.
\section{Related Work}
\label{label_related_work}

In the context of benchmarking inference accelerators, MLPerf~\cite{libutti_igul} evaluates inference time and energy consumption. Gopinath et al.~\cite{gopinath_ravisankar} assess methods for data storage and transmission. Liang et al.~\cite{liang_zhang_lu} proposed a survey addressing the challenges of communication overhead in large-scale distributed ML, with the goal of enhancing communication efficiency. Their work covers topics such as model synchronization, communication data compression, resource allocation, task scheduling, and modern communication infrastructures. Similarly, Rebai et al.~\cite{rebai_ojewale_ullah} introduced an approach to reduce communication overhead in distributed ML by incorporating compression directly into the network interface card, thereby aiming to improve overall training efficiency.

Rijnsdorp et al.~\cite{rijnsdorp_zwol} proposed a GNSS interference detection and classification module utilizing commercial off-the-shelf components. Their module employs signal spectrum analysis techniques, such as kurtosis and power spectral density calculations, combined with a ML model to detect and classify anomalies in incoming signals. Ghobadi et al.~\cite{ghobadi_spogli} accurately separated refractive and diffractive effects in GNSS signals by employing the fast iterative filtering technique. This method decomposes non-stationary GNSS raw phase data into intrinsic mode components, allowing for precise identification of the cutoff frequency for phase detrending. Li et al.~\cite{li_huang_lang} proposed an Twin SVM approach for GNSS interference classification, which achieves faster processing times. However, the method is sensitive to the selection of hyperparameters and the quality of input features, potentially affecting its generalization across diverse interference scenarios. Reda et al.~\cite{reda_mekkawy} combined attention mechanisms with mutual information-based feature selection to identify the most relevant features for distinguishing between normal and jammed GNSS signals. This technique aims to improve detection accuracy while reducing computational complexity. While the approach enhances the understanding of ionospheric impacts, it is not specifically designed to classify intentional interference sources. Moreover, the reliance on high-resolution phase data and complex signal decomposition may present challenges for real-time interference classification. In summary, the approaches encounter challenges in processing large-scale datasets due to the increased computational complexity and the need for further data compression. 

Rügamer et al.~\cite{ruegamer_lukcin} proposed a compressed sensing random demodulator AIC architecture for interference detection, enabling efficient monitoring of GNSS bands, with a bandwidth of up to $100\,\text{MHz}$. Li et al.~\cite{li_wang_zhong} transforms GNSS signals from the time to the frequency domain using the discrete cosine transform (DCT) and frequency division, retaining crucial information. This approach aims to reduce data volume while preserving essential signal characteristics for accurate carrier-to-noise density ($C/N_0$) estimation. Rakhmanov et al.~\cite{rakhmanov_wiseman} efficiently transmit GNSS information to autonomous vehicles through data compression. Dakic et al.~\cite{dakic_homssi} presented a framework that utilizes spiking neural networks (SNNs) for detecting IoT signals amidst uplink interference in low Earth orbit (LEO) satellite constellations. The study highlights the ultra-low power consumption of SNNs, making them particularly suitable for onboard signal detection in IoT LEO satellites. However, training SNNs is more complex compared to traditional ML models, and ensuring that SNNs generalize effectively across diverse interference scenarios while maintaining robustness in real-world conditions remains a challenge. However, existing methods still face challenges in balancing compression efficiency with signal interpretability and generalization across diverse interference scenarios. This motivates the exploration of disentangled representations using VAEs and GANs~\cite{iqbal_aman_sikdar,mascher_laller}, which can enable more effective compression while preserving semantically meaningful features for robust GNSS interference monitoring.
\section{Methodology}
\label{label_method}

We begin by presenting our method based on the factorized VAE (FactorVAE) for disentanglement, which is employed for data compression (Section~\ref{label_method_factor_compr}). Next, we introduce the vanilla VAE for interpolation-based data augmentation (Section~\ref{label_method_vae_augm}), which is also applied to FactorVAE (Section~\ref{label_method_factor_augmen}) and CVAE-GAN (Section~\ref{label_method_cvae_gan_augm}).

\subsection{Factorized VAE for Compressed Disentanglement}
\label{label_method_factor_compr}

To achieve spectrogram compression, we employ disentanglement, a process that involves decomposing complex data into distinct, independent factors or components, where each factor represents a specific underlying structure or feature. The primary objective is to extract meaningful data representations. A disentangled representation ensures that each latent variable corresponds to a single underlying factor. Given the assumption that these factors vary independently, the target is to achieve a factorial distribution. The $\beta$-VAE~\cite{higgins_matthey} is a widely used method for unsupervised disentanglement, extending the VAE framework proposed by Kingma \& Welling~\cite{kingma_welling} for generative modeling. This approach modifies the standard VAE objective by introducing an increased weight ($\beta > 1$) on the Kullback-Leibler (KL) divergence between the variational posterior and the prior distribution. However, in $\beta$-VAE, an improved level of disentanglement often comes at the cost of reconstruction quality. The challenge, therefore, lies in optimizing the trade-off between disentanglement and reconstruction, with the aim of enhancing disentanglement while minimizing reconstruction loss.

Our approach to compressed sensing leverages FactorVAE~\cite{kim_mnih} to extract essential features from GNSS spectrograms within a lower-dimensional space, utilizing only these features for classification. FactorVAE extends the standard VAE framework by introducing an additional penalty term that promotes a factorial structure in the marginal distribution of latent representations, while preserving reconstruction quality. Formally, we define observations $X^{(i)} \in \mathcal{X}$, where $i \in \{1, \ldots, N\}$ as being generated by a combination of $K$ underlying factors $f = (f_1, \ldots, F_K)$. These observations are modeled using a real-valued latent vector $z \in \mathbb{R}^d$, which serves as a compact representation of the data. The generative model assumes a standard Gaussian prior over the latent space, defined as $p(z) = \mathcal{N}(0, I)$, ensuring a factorized distribution. The decoder, denoted as $p_{\theta}(x|z)$, is parameterized by a neural network. The variational posterior for an observation is expressed as $q_{\theta}(z|x) = \prod_{j=1}^d \mathcal{N} \big(Z_j|\mu_j(x), \sigma_j^2(x) \big)$, where the mean $\mu_j(x)$ and variance $\sigma_j^2(x)$ are produced by the encoder network. This variational posterior represents the distribution of the latent representation corresponding to a given data point $x$~\cite{kim_mnih}. Consequently, the distribution of representations for the entire dataset is given by
\begin{equation}
\label{equ_vae1}
    q(z) = \mathbb{E}_{p_\text{data}} \big[q(z|x) \big]= \frac{1}{N} \sum_{i=1}^Nq(z|x^{(i)}),
\end{equation}
referred to as the marginal posterior. In a disentangled representation, each latent variable $z_j$ corresponds uniquely to a single underlying factor $f_k$, ensuring structure in the latent space, reflected in the factorized distribution $q(z) = \prod_{j=1}^d q(z_j)$. The objective function of the $\beta$-VAE is given by
\begin{equation}
\label{equ_vae2}
    \frac{1}{N}\sum_{i=1}^N \big[\mathbb{E}_{q(z|x^{(i)})}[\log p(x^{(i)}|z)] - \beta \text{KL} \big(q(z|x^{(i)}) || p(z)\big) \big],
\end{equation}
that serves as a variational lower bound on $\mathbb{E}_{p_{\text{data}}(x)} [\log p(x^{(i)})]$ for $\beta \geq 1$. The first term corresponds to the negative reconstruction error, while the second term represents a complexity penalty that functions as a regularizer. Given
\begin{equation}
\label{equ_vae3}
    \mathbb{E}_{p_{\text{data}}(x)} \big[\text{KL}\big(q(z|x) || p(z)\big)\big] = I(x;z) + \text{KL}(q(z) || p(z)),
\end{equation}
where $I(x;z)$ denotes the mutual information between $x$ and $z$ under the joint distribution $p_{\text{data}}(x)q(z|x)$. Increasing $\beta > 1$ imposes a stronger penalty on both terms, promoting greater disentanglement at the cost of reduced reconstruction quality. If this reduction becomes too severe, the latent variables fail to retain sufficient information about the observations, rendering the recovery of the true factors infeasible. Hence, FactorVAE enhances the VAE objective by
\begin{equation}
\label{equ_vae4}
\begin{split}
    \frac{1}{N} \sum_{i=1}^N \big[\mathbb{E}_{q(z|x^{(i)})}[\log p(x^{(i)} | z)] &- \text{KL} \big(q(z|x^{(i)})||p(z) \big)\big] \\&- \gamma \text{KL} \big(q(z)||\overline{q}(z) \big),
\end{split}
\end{equation}
where $\overline{q}(z) := \prod_{j=1}^d q(z_j)$, representing the Total Correlation (TC). In FactorVAE, a batch of samples is drawn  from $q(z)$, and each latent dimension is randomly permuted across the batch. Provided that the batch size is sufficiently large (in our case, 64), the distribution of these permuted samples serves as a close approximation of $\overline{q}(z)$~\cite{kim_mnih}.

\begin{figure}[!t]
    \centering
    \includegraphics[width=1.0\linewidth]{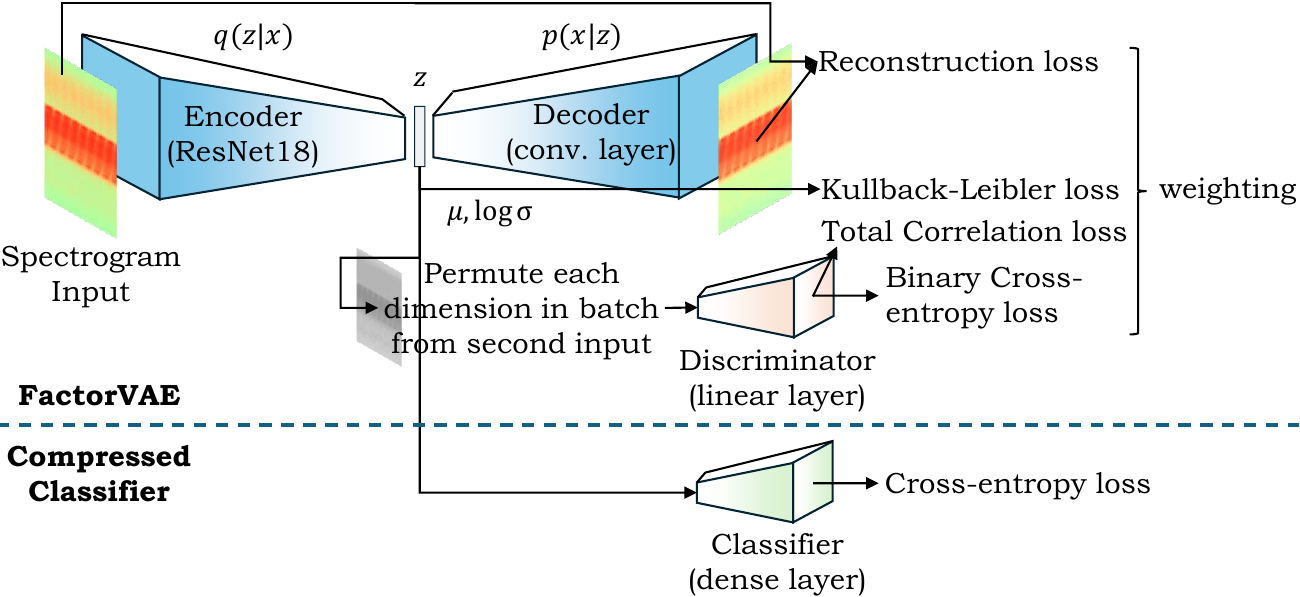}
    \caption{Overview of the FactorVAE method for compressed GNSS interference disentanglement.}
    \label{figure_factorvae}
\end{figure}

Figure~\ref{figure_factorvae} presents an overview of our proposed pipeline. The input consists of a GNSS spectrogram, with dimensions varying depending on the dataset, taking values of $1024 \times 32$, $1024 \times 256$, or $512 \times 256$. The spectrogram is processed through an encoder architecture, where we evaluate ResNet18 and convolutional blocks\footnote{The \textit{encoder} is the non-transposed decoder.} for feature extraction. The encoder outputs a latent representation $z \in \mathbb{R}^d$ with dimensionality $d$, where we assess values $d \in \{16, 32, 64, 128, 256\}$. The decoder\footnote{The \textit{decoder} architecture consists of five decoder blocks, each incorporating convolutional Transpose2D layers with progressively increasing output sizes. Each layer employs a kernel size of 5, a padding of 2, a stride of 2, and an output padding of 1. Additionally, batch normalization with a momentum of 0.9 is applied following each convolutional layer.} reconstructs the original spectrogram, optimizing the MSE as the reconstruction loss. A discriminator\footnote{The \textit{discriminator} comprises six fully connected linear layers, each containing 1,000 units. A LeakyReLU activation function with a negative slope coefficient of 0.2 is applied to each layer.} is trained on the latent space to distinguish between real and generated spectrograms using the cross-entropy (CE) loss. The total loss function is defined as the sum of the reconstruction loss and the KL divergence loss. The VAE encoder and decoder are optimized using the Adam optimizer with predefined hyperparameters, and training is conducted over 250 epochs. After training, the latent variables are extracted and used to train a classification network composed of one, two, or three fully connected layers. Batch normalization (BN) and ReLU activation functions are optionally incorporated between layers. An extensive hyperparameter search is performed across all model parameters. Finally, we compare training and inference times, as well as compression rates, between two models: (1) a baseline \textit{ResNet18} model trained with CE loss, and (2) a model consisting of \textit{dense} layers trained exclusively on the extracted latent variables using CE loss. We conduct a comparison of the compression performance between the two models across the controlled small-scale and large-scale indoor dataset and real-world highway dataset 1 presented in Section~\ref{label_experiments}.

\subsection{VAE for Data Augmentation}
\label{label_method_vae_augm}

\begin{figure}[!t]
    \centering
    \includegraphics[width=1.0\linewidth]{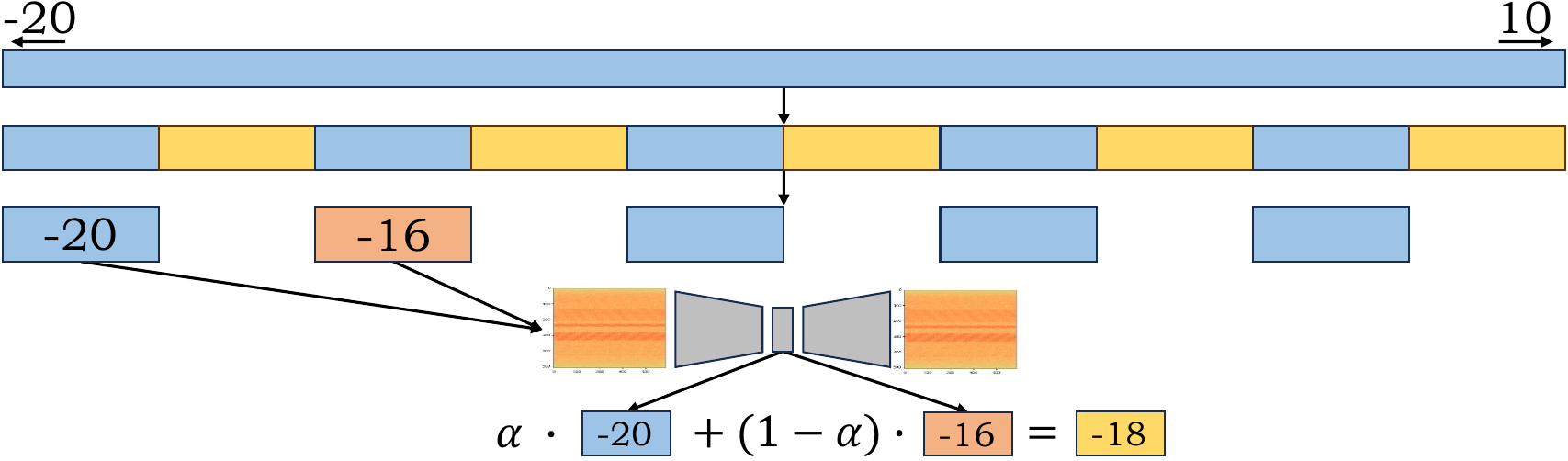}
    \caption{Visualization of the interpolation of spectrograms between two signal power levels, -20 and -16, as a function of the interpolation parameter $\alpha$.}
    \label{figure_snr_interpolation}
\end{figure}

By independently manipulating latent factors, we can generate augmented samples that preserve key structural characteristics while introducing controlled variations, thereby improving model performance in scenarios with limited training data. The primary goal is to synthesize new spectrograms for use in classification tasks. Building on the VAE framework proposed by Kingma \& Welling~\cite{kingma_welling}, we extract general features applicable to various tasks. Figure~\ref{figure_snr_interpolation} illustrates the interpolation between two spectrograms. Specifically, we extract latent features from two distinct spectrograms, interpolate these features based on the interpolation parameter $\alpha$, and then reconstruct the interpolated spectrograms. Furthermore, we interpolate the signal power within the range of -20 to +10, skipping every second bin during the interpolation process. This procedure is applied individually to each jammer subclass. The objective is to train a generalized model capable of accurately classifying all signal power levels. We conduct experiments on the controlled large-scale indoor dataset (see Section~\ref{label_experiments}).

\subsection{Factorized VAE for Data Augmentation}
\label{label_method_factor_augmen}

We apply the same data augmentation technique for FactorVAE~\cite{kim_mnih}, as outlined in Section~\ref{label_method_factor_compr}, to interpolate the signal powers between two spectrograms.

\begin{figure}[!t]
    \centering
    \includegraphics[width=1.0\linewidth]{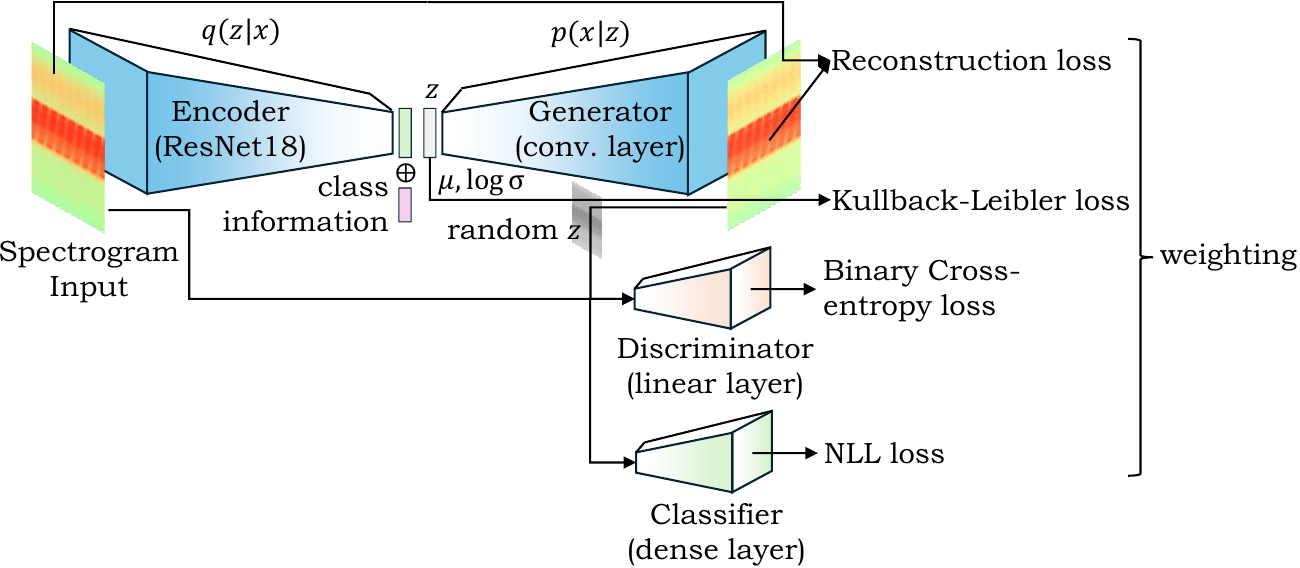}
    \caption{Visualization of our CVAE-GAN variant.}
    \label{figure_cvae_gan}
\end{figure}

\subsection{CVAE-GAN for Data Augmentation}
\label{label_method_cvae_gan_augm}

Figure~\ref{figure_cvae_gan} presents an overview of the CVAE-GAN architecture~\cite{bao_chen_wen} employed for data augmentation. The input spectrogram is first processed by a ResNet18-based encoder model. Class information (i.e., the conditioning variable) is integrated via concatenation in the final layer of this encoder, followed by additional linear layers. The latent variable $z_j$, reparameterized from the mean ($\mu$) and variance ($\sigma$) values, is utilized by the generator to reconstruct input spectrograms while minimizing the MSE loss. Additionally, a randomly sampled latent variable is fed into the generator to train the discriminator\footnote{The \textit{discriminator} consists of a sequence of seven convolutional layers followed by a fully connected linear layer.} using the binary CE loss. To optimize both the encoder and the generator, we employ mean feature matching as described by Bao et al.~\cite{bao_chen_wen}. The classifier is implemented as a ResNet18 model, trained using the negative log-likelihood (NLL) loss. All loss functions are appropriately weighted. Finally, we interpolate signal power levels in the controlled small-scale indoor dataset and vary the distances between the jamming source and the receiver in the highway dataset 2.
\section{Experiments}
\label{label_experiments}

\begin{figure*}[!t]\captionsetup[subfigure]{font=footnotesize}
    \centering
	\begin{minipage}[t]{0.195\linewidth}
        \centering
        \includegraphics[trim=40 40 40 60, clip, width=1.0\linewidth]{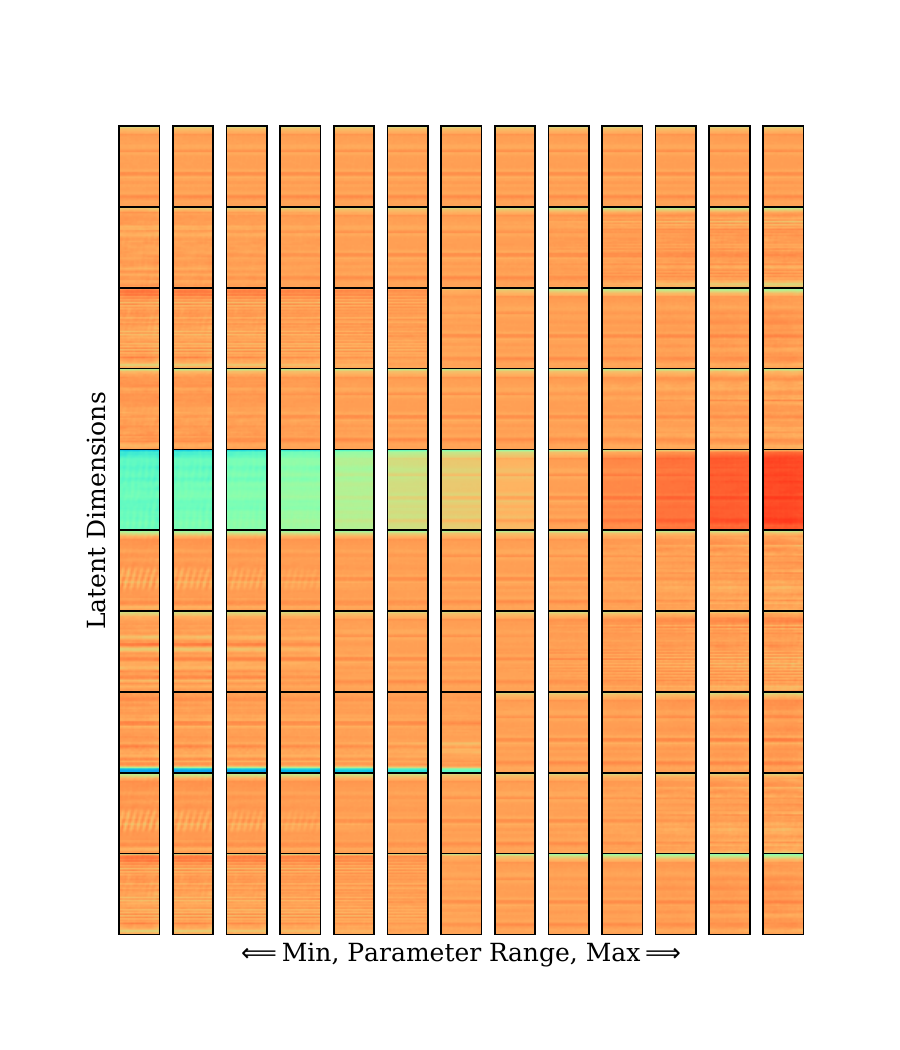}
        \subcaption{FactorVAE on the indoor small-scale dataset.}
        \label{figure_latent_vectors1}
    \end{minipage}
    \hfill
	\begin{minipage}[t]{0.292\linewidth}
        \centering
        \includegraphics[trim=64 42 62 62, clip, width=1.0\linewidth]{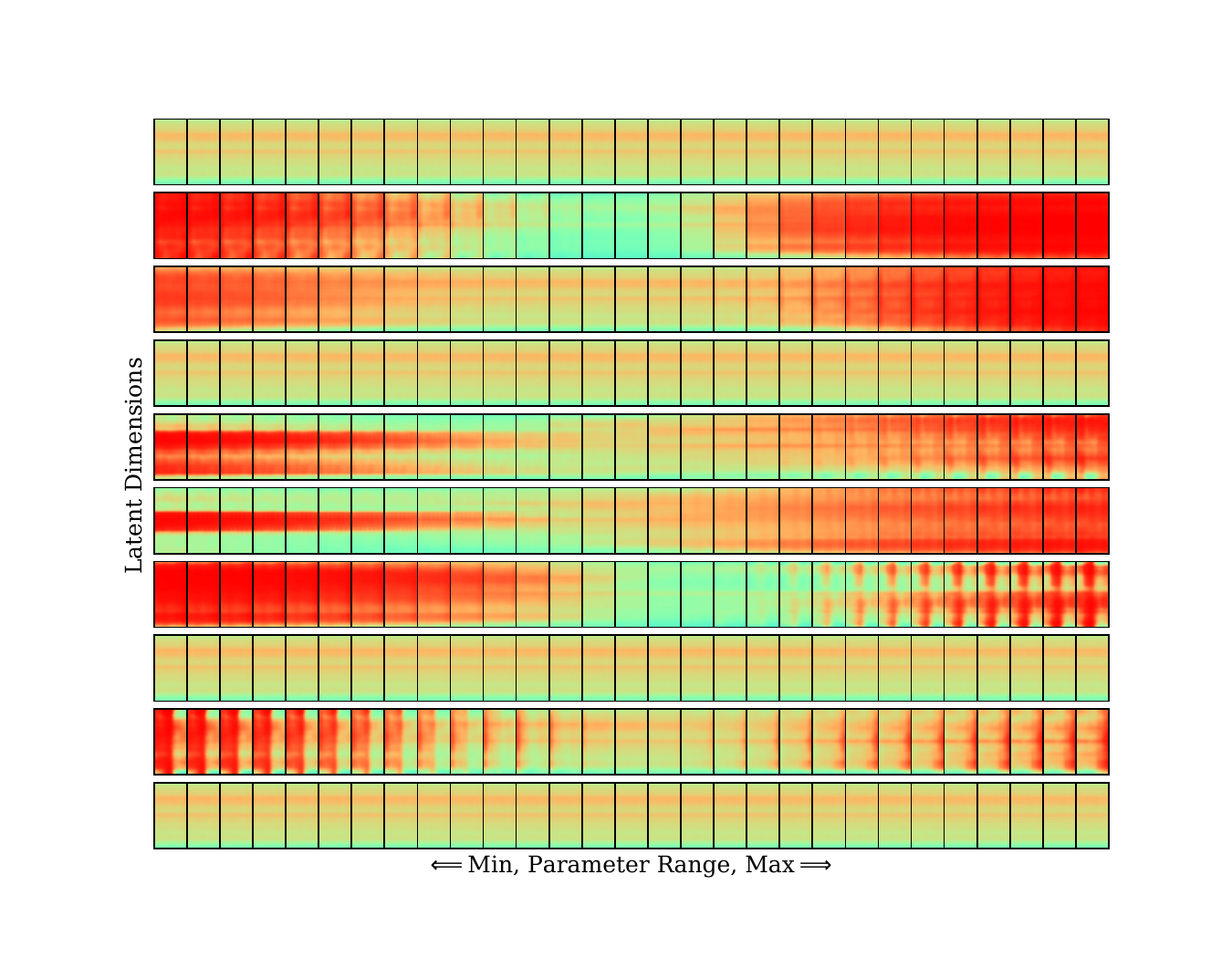}
        \subcaption{FactorVAE on the real-world highway dataset 1.}
        \label{figure_latent_vectors2}
    \end{minipage}
    \hfill
	\begin{minipage}[t]{0.157\linewidth}
        \centering
        \includegraphics[trim=60 78 52 78, clip, width=1.0\linewidth]{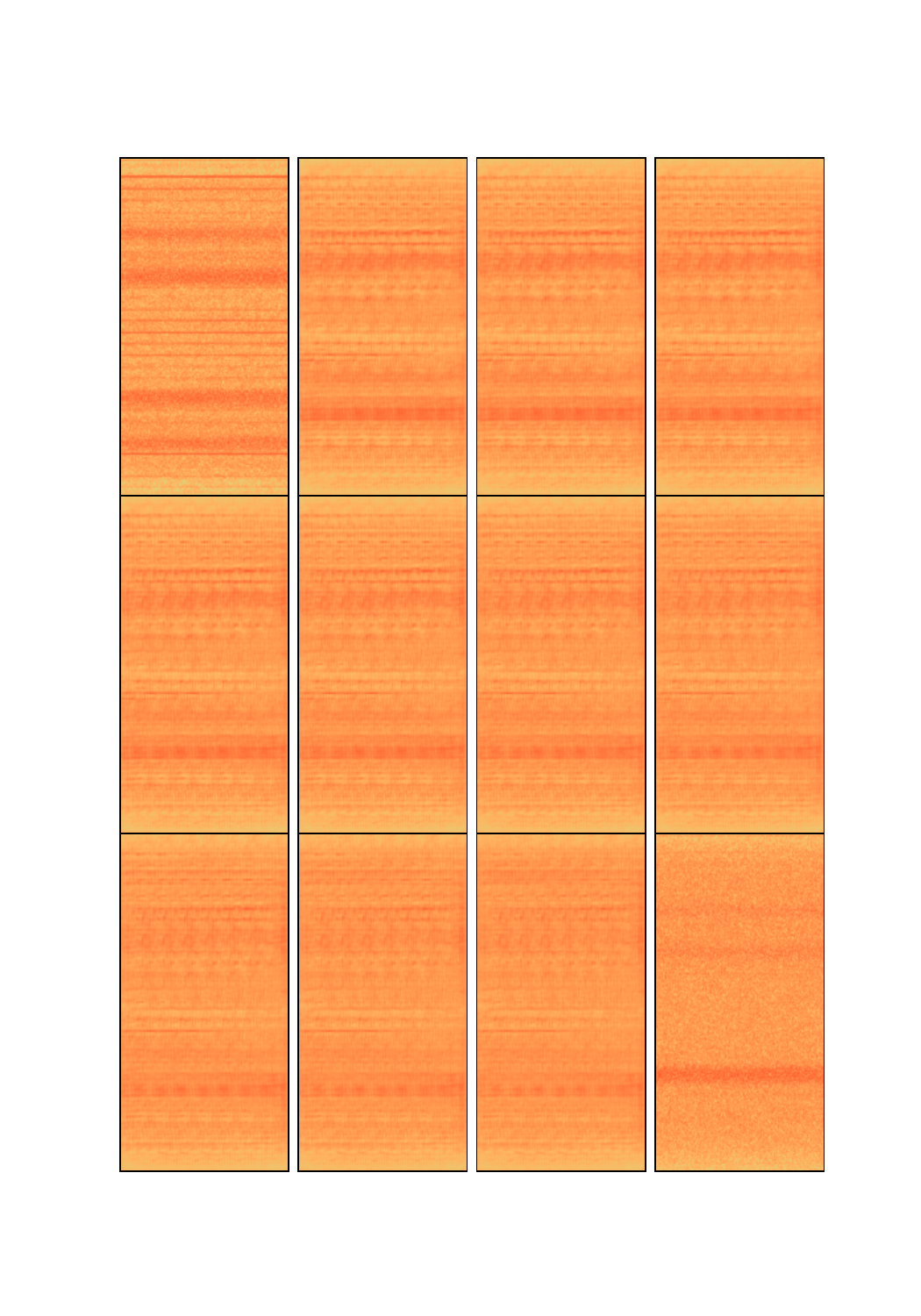}
        \subcaption{Signal powers with CVAE-GAN on the highway dataset 1.}
        \label{figure_latent_vectors3}
    \end{minipage}
    \hfill
	\begin{minipage}[t]{0.157\linewidth}
        \centering
        \includegraphics[trim=60 78 52 78, clip, width=1.0\linewidth]{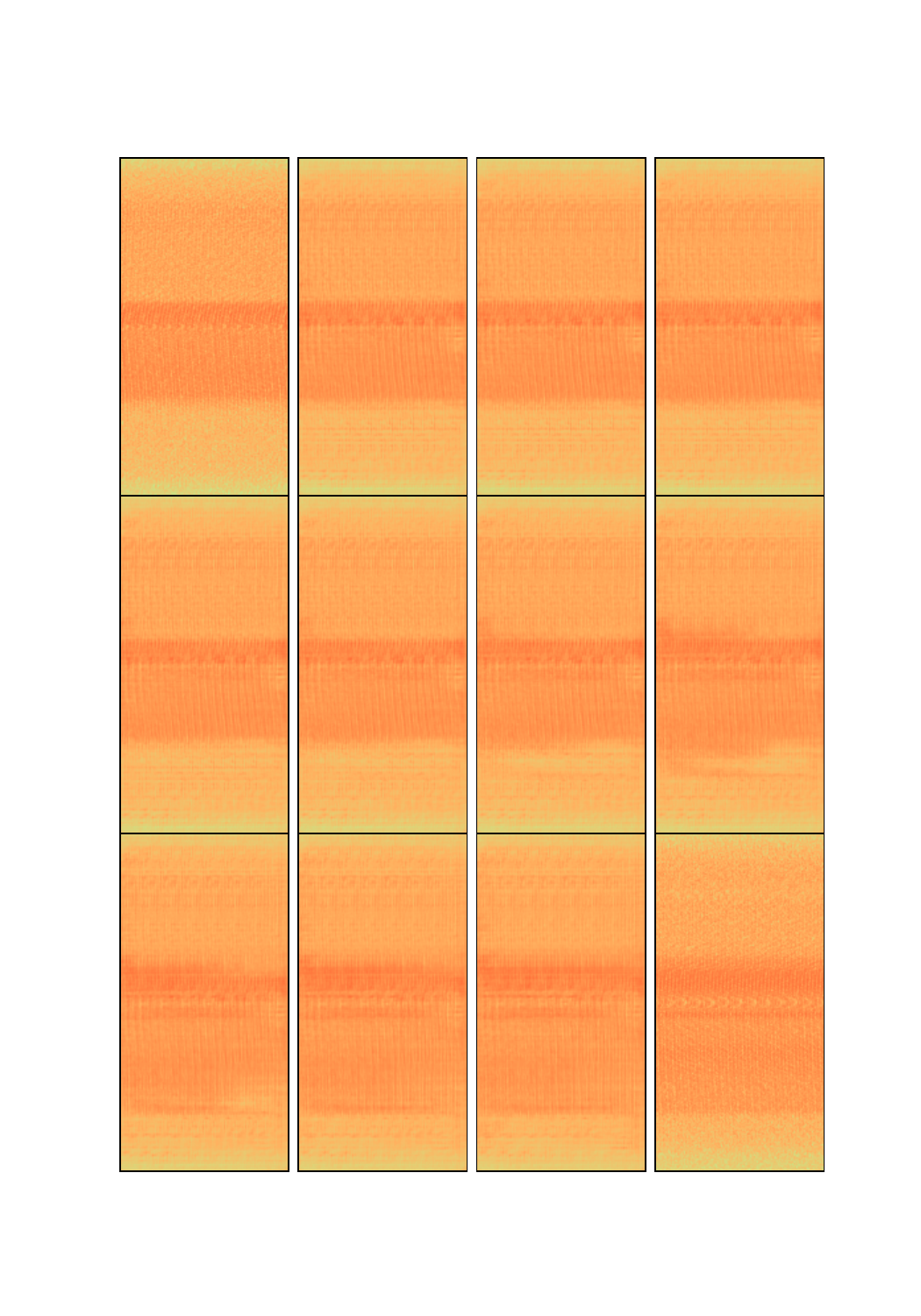}
        \subcaption{Bandwidths with CVAE-GAN on the highway dataset 1.}
        \label{figure_latent_vectors4}
    \end{minipage}
    \hfill
	\begin{minipage}[t]{0.157\linewidth}
        \centering
        \includegraphics[trim=60 78 52 78, clip, width=1.0\linewidth]{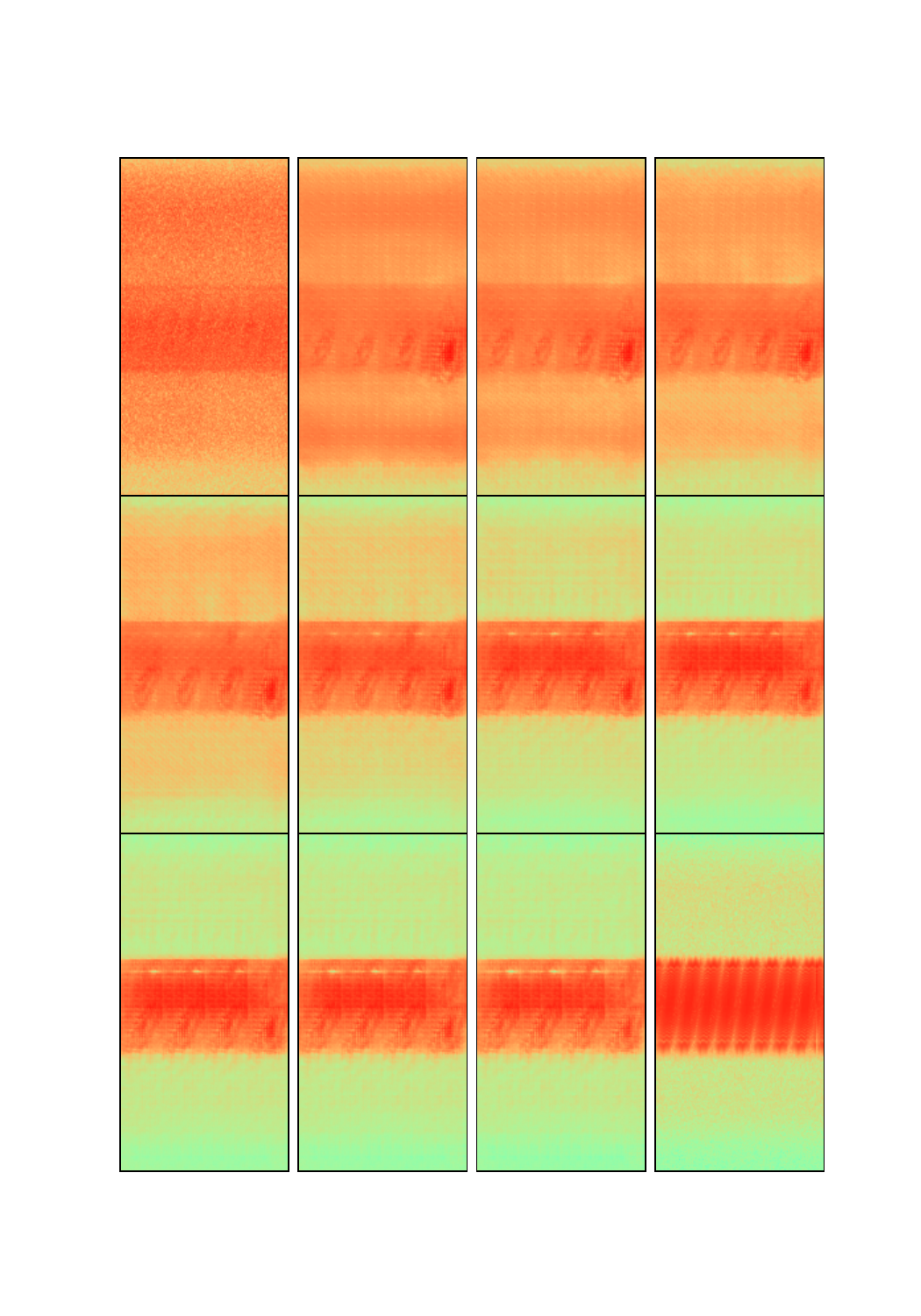}
        \subcaption{Distances with CVAE-GAN on the highway dataset 2.}
        \label{figure_latent_vectors5}
    \end{minipage}
    \vspace{-0.1cm}
    \caption{Visualization of latent traversal (a and b) and interpolation between spectrograms for data augmentation (c to e).}
    \label{figure_latent_vectors}
    \vspace{-0.3cm}
\end{figure*}

To thoroughly evaluate compression and data augmentation techniques with disentanglement, multiple datasets representing diverse conditions are essential. Using varied data sources provides a comprehensive understanding of performance across scenarios, assessing factors like compression rate, inference time, and their correlation with data input. Incorporating variations such as signal power interpolation, bandwidths, and receiver-jammer distance enables a robust analysis of performance and scalability. Evaluating two indoor and two outdoor real-world datasets ensures accurate, representative assessments of complex environments. The data collection for the \textbf{controlled small-scale indoor dataset}~\cite{brieger_ion_gnss} took place at the Fraunhofer IIS L.I.N.K.~test center, using six low-cost GNSS sensors in a controlled, interference-free environment to study GNSS jammers. Over two days, various multipath scenarios were tested by altering barriers and jammer positions, capturing 33 interference signal types across seven waveform classes. The dataset includes 5-minute recordings per scenario, featuring both clean and jammed signals at different attenuation levels to analyze varying interference intensities. Furthermore, we employ the \textbf{controlled large-scale indoor dataset}~\cite{heublein_feigl_crpa}, which was recorded at the L.I.N.K.~center. The experimental setup consisted of a receiver antenna -- specifically, a patch antenna array comprising four identical coaxial-fed square patch elements -- positioned at one end of the hall, while an MXG signal generator was placed at the opposite end. Data acquisition was performed under various configurations, including scenarios in an unoccupied environment as well as setups incorporating absorber walls between the antenna and the generator. This setup facilitated the capture of distinct multipath effects and cases of significant signal absorption. The dataset enables interpolation across different bandwidths and signal power levels. For the \textbf{real-world highway dataset 1}~\cite{ott_heublein_icl}, a sensor station was deployed on a bridge along a highway to capture short, wideband snapshots within the E1 and E6 GNSS frequency bands. The system recorded raw IQ snapshots with a duration of $20\,ms$, operating at a sampling rate of $62.5\,\text{MHz}$, an analog bandwidth of $50\,\text{MHz}$, and an 8-bit resolution. Data streams were manually analyzed by experts, resulting in the classification of snapshots into 11 distinct categories. We next utilize the \textbf{real-world highway dataset 2}~\cite{heublein_raichur_ion}, which involves the placement of two sensor stations adjacent to the highway at the \textit{Test Bed Lower Saxony}, located near Braunschweig. The dataset comprises manual selection of 21 recordings, incorporating \textit{chirp} interferences. The intensity of the interference diminishes as the distance between the vehicle and the sensor stations increases, while it intensifies when the vehicle passes directly next to the stations. As a result, both the distance and driving direction of the vehicle can be accurately predicted.
\section{Evaluation}
\label{label_evaluation}

\begin{figure}[!t]
    \centering
    \includegraphics[trim=107 48 86 68, clip, width=1.0\linewidth]{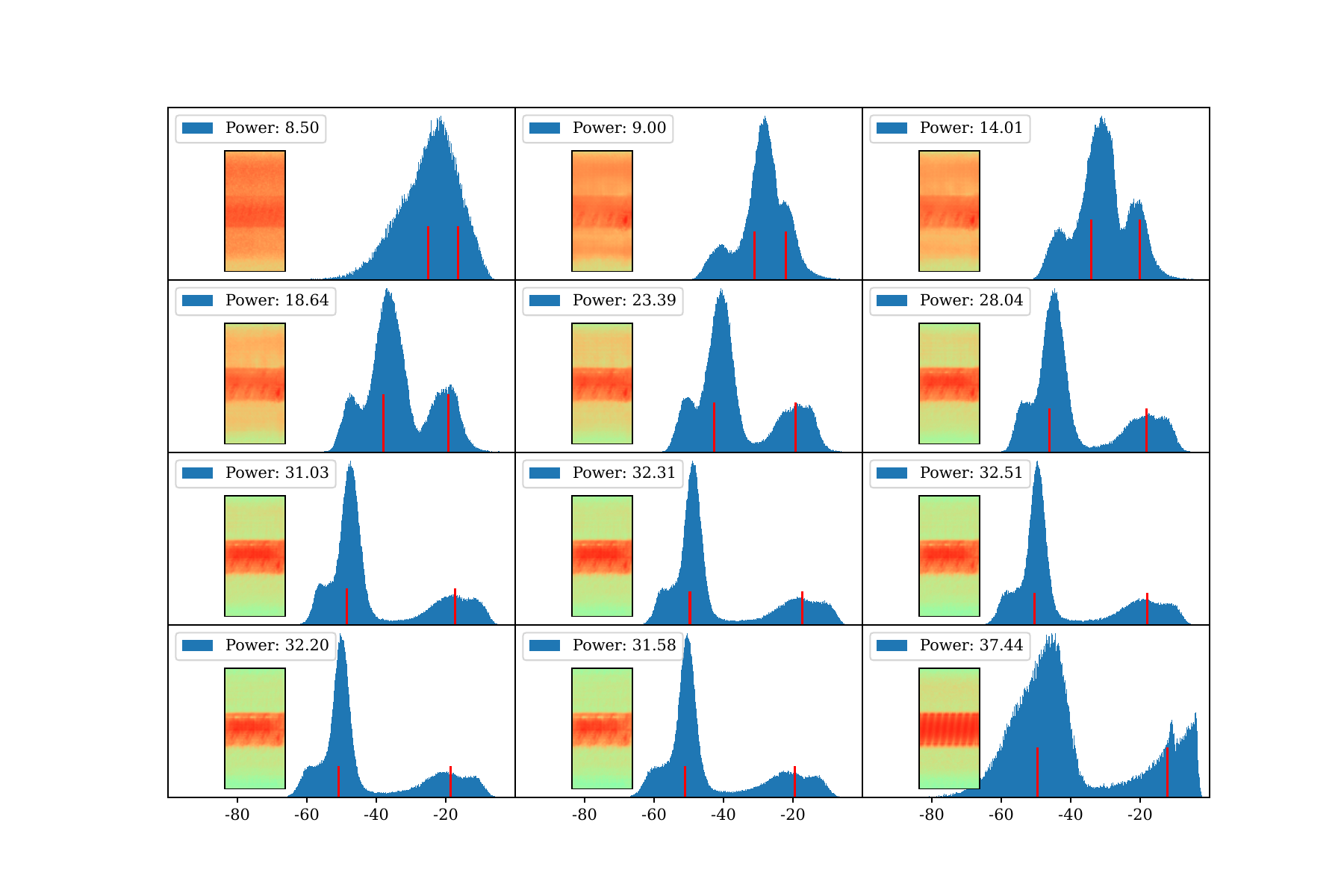}
    \caption{Histogram plots of the signal power are presented for each interpolated spectrogram. The red lines indicate the mean signal power levels corresponding to noise (left line) and interference (right line). The legend represents the difference between these two mean values.}
    \label{figure_histograms}
\end{figure}

\begin{figure*}[!t]\captionsetup[subfigure]{font=footnotesize}
    \centering
	\begin{minipage}[t]{0.245\linewidth}
        \centering
        \includegraphics[trim=9 10 10 10, clip, width=1.0\linewidth]{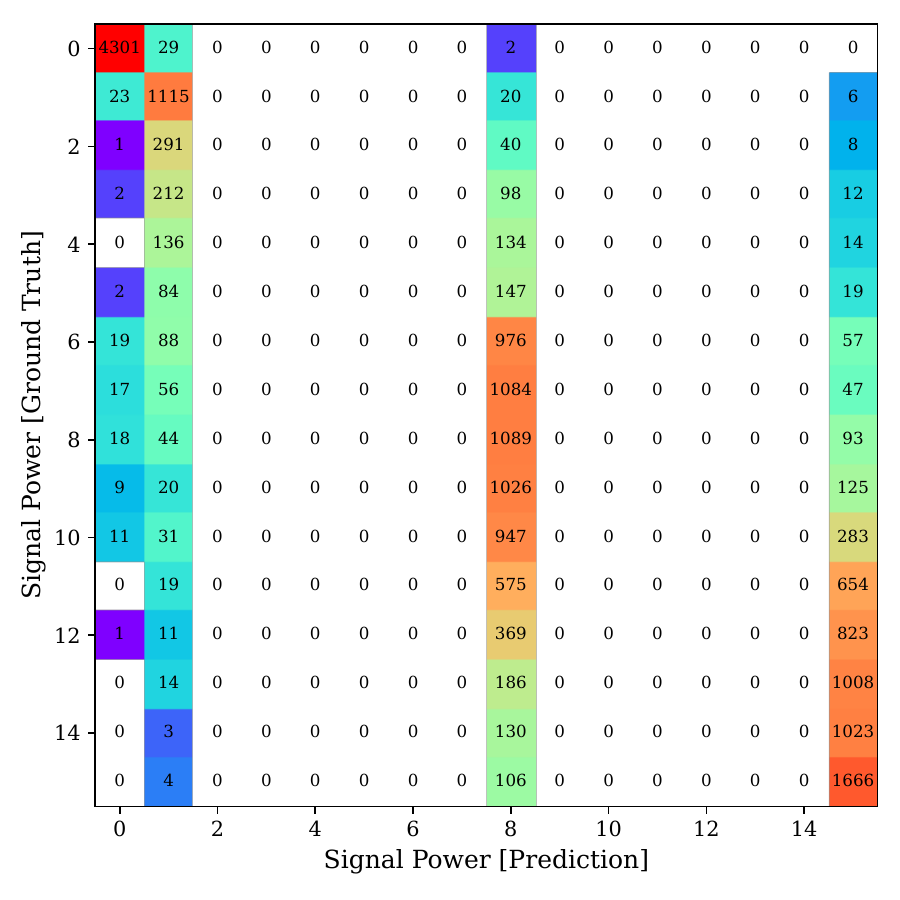}
        \subcaption{Without interpolation.}
        \label{figure_conf_matrix1}
    \end{minipage}
    \hfill
	\begin{minipage}[t]{0.245\linewidth}
        \centering
        \includegraphics[trim=9 10 10 10, clip, width=1.0\linewidth]{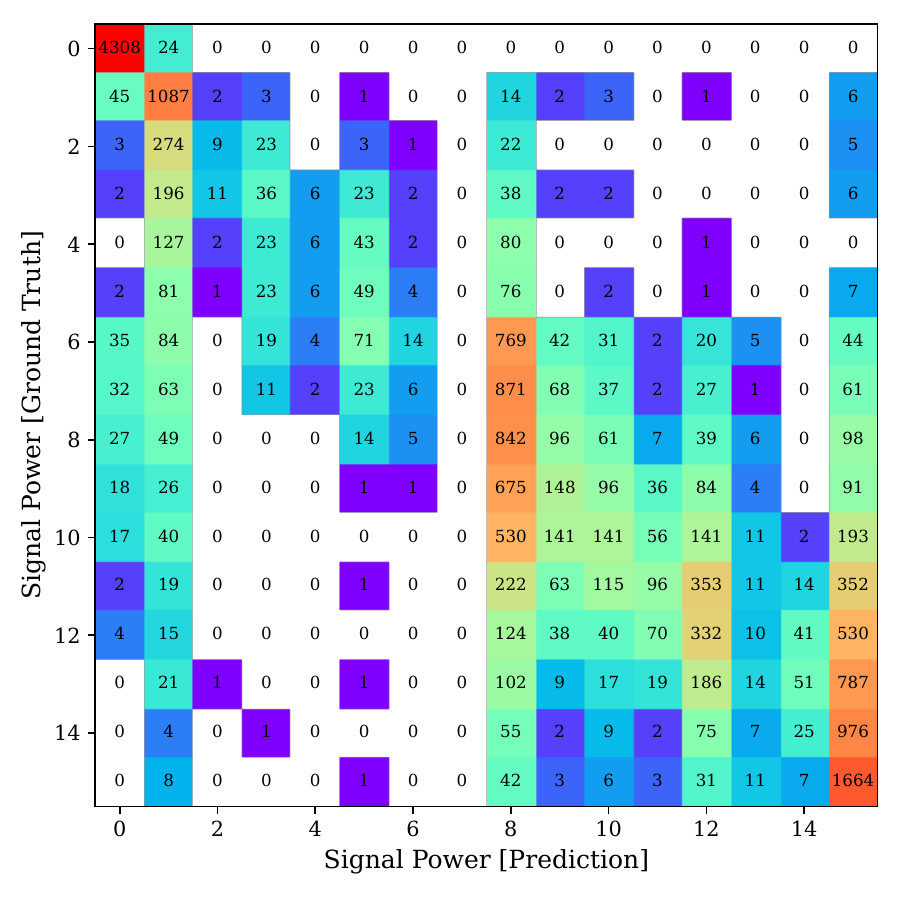}
        \subcaption{With interpolation.}
        \label{figure_conf_matrix2}
    \end{minipage}
    \hfill
	\begin{minipage}[t]{0.245\linewidth}
        \centering
        \includegraphics[trim=9 10 10 10, clip, width=1.0\linewidth]{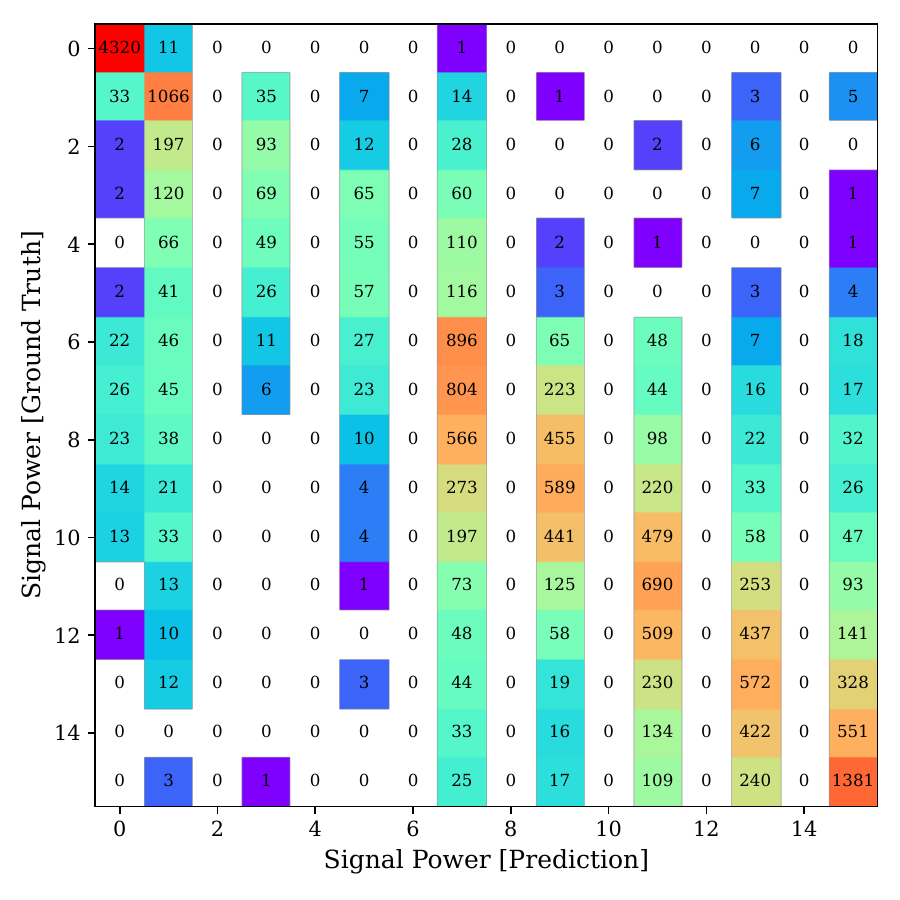}
        \subcaption{Without interpolation.}
        \label{figure_conf_matrix3}
    \end{minipage}
    \hfill
	\begin{minipage}[t]{0.245\linewidth}
        \centering
        \includegraphics[trim=9 10 10 10, clip, width=1.0\linewidth]{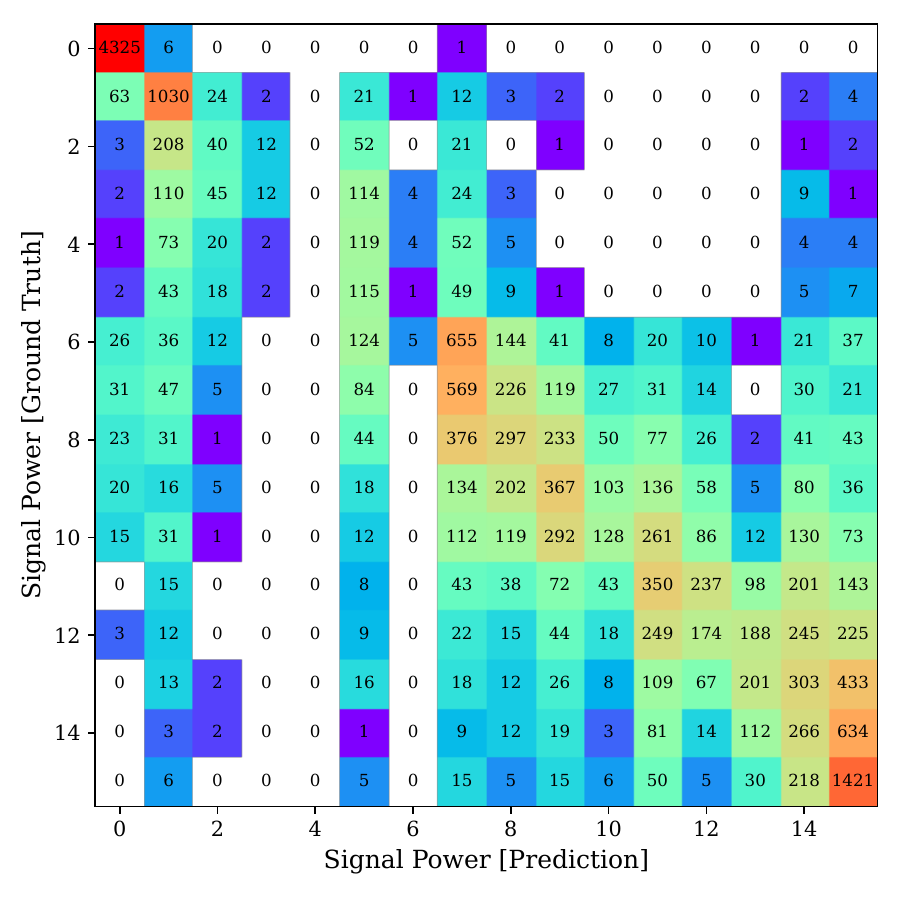}
        \subcaption{With interpolation.}
        \label{figure_conf_matrix4}
    \end{minipage}
    \vspace{-0.1cm}
    \caption{Confusion matrices of the ResNet18 model based on a subset or interpolated values of the training set. Subfigures (a) and (b) present evaluations using four distinct bins, whereas subfigures (c) and (d) depict evaluations conducted with nine bins.}
    \label{figure_conf_matrix}
    \vspace{-0.4cm}
\end{figure*}

\textbf{Ablation Study.} Figure~\ref{figure_latent_vectors} illustrates latent traversal and interpolation across signal powers, bandwidths, and distances for various datasets using FactorVAE and CVAE-GAN. In the disentanglement analysis of the indoor small-scale dataset (Figure~\ref{figure_latent_vectors1}), different latent dimensions influence distinct interference classes -- for instance, rows 6 and 9 affect \textit{chirp}, while row 5 influences background \textit{noise} in the absence of interference. In the real-world highway dataset 1 (Figure~\ref{figure_latent_vectors2}), FactorVAE introduces artifacts but successfully disentangles \textit{chirp} interference. CVAE-GAN effectively interpolates different signal power levels (Figure~\ref{figure_latent_vectors3}) but struggles to achieve smooth transitions between bandwidths (Figure~\ref{figure_latent_vectors4}), while successfully interpolating across different distances (Figure~\ref{figure_latent_vectors5}). Figure~\ref{figure_histograms} further examines the interpolation of the interference-to-noise ratio (INR), defined as the ratio between interference and background noise. The histogram plots depict the corresponding spectrograms on the indoor large-scale dataset, illustrating interpolation from the top-left to the bottom-right spectrograms. These distributions, separated by interference power and noise, highlight their respective averages with red lines. The results indicate that as INR increases, the distributions become more distinct, demonstrating that CVAE-GAN effectively interpolates signal powers.

\textbf{Data Augmentation Evaluation.} Figure~\ref{figure_conf_matrix} presents the evaluation results for data augmentation on the indoor large-scale dataset. The ResNet18 model is trained exclusively on three bins in Figure~\ref{figure_conf_matrix1} (comprising non-interference and two signal power levels) and on nine bins in Figure~\ref{figure_conf_matrix3} (non-interference and eight signal power levels). Intermediate bins are interpolated using FactorVAE in Figure~\ref{figure_conf_matrix2} and Figure~\ref{figure_conf_matrix4}, followed by training on the augmented data with ResNet18. The results indicate that without interpolation, the model can only predict previously learned signal power levels, whereas with data augmentation, it successfully classifies intermediate values, albeit with minor confusion among adjacent bins. Data augmentation using FactorVAE exhibits superior performance.

\begin{figure}[!t]
    \centering
	\begin{minipage}[t]{1.0\linewidth}
        \centering
        \includegraphics[trim=10 10 10 10, clip, width=1.0\linewidth]{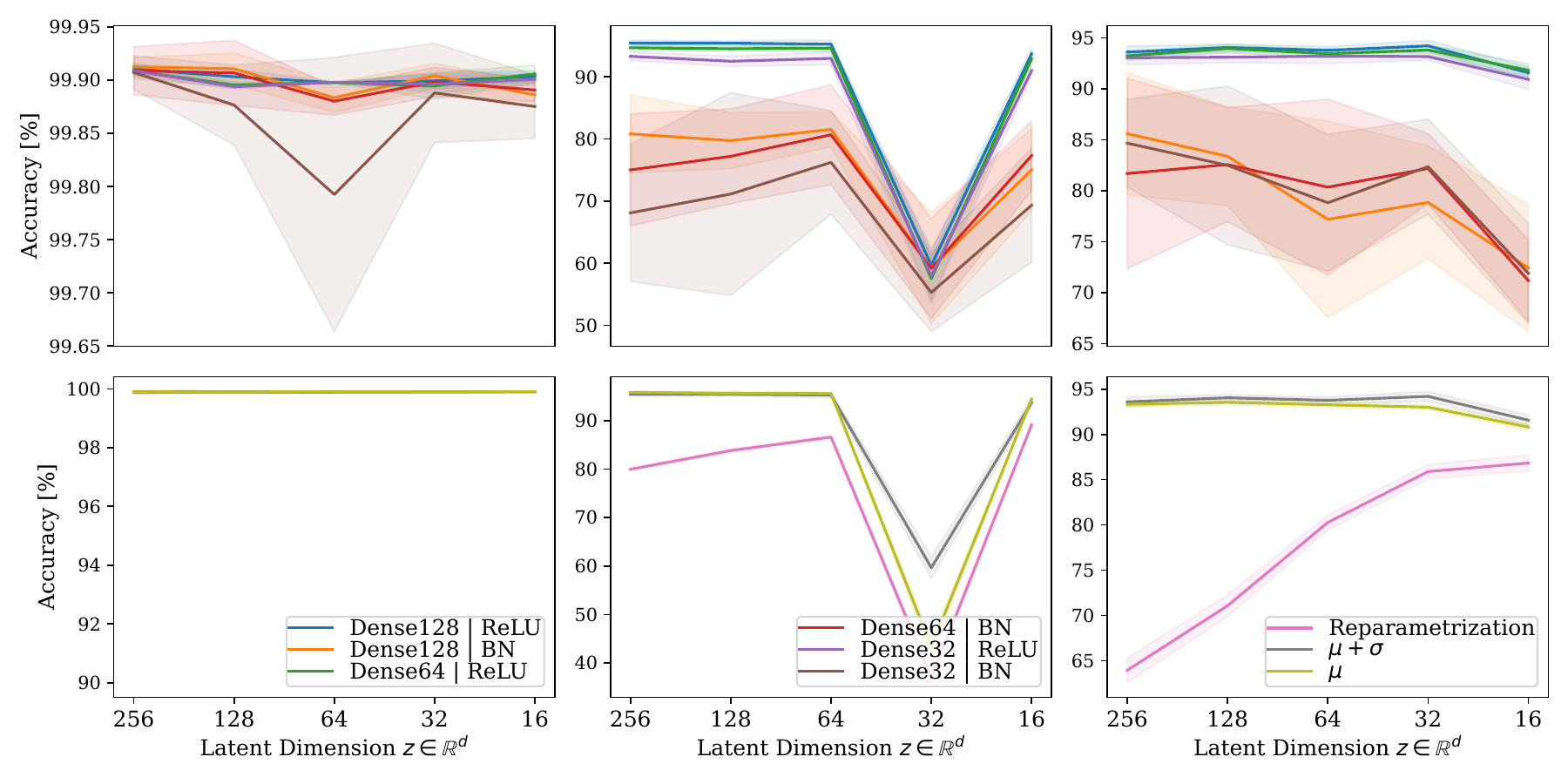}
    \end{minipage}
    \vspace{-0.4cm}
    \caption{Hyperparameter searches of the number of dense layers, ReLU activation, and BN for the compressed classification model (top), and of $\mu$, $\sigma$, and reparametrization for FactorVAE (bottom). Left: real-world highway dataset 1. Middle: indoor large-scale dataset. Right: indoor small-scale dataset.}
    \label{figure_hyperparameter}
\end{figure}

\textbf{Hyperparameter Searches.} Figure~\ref{figure_hyperparameter} illustrates the hyperparameter search across three datasets, focusing on the dimensionality of the latent variable $z$, the number of dense layers, and the inclusion of ReLU activation and BN, as shown in the top row. Subsequently, the best-performing configuration (i.e., dense layers of size 128 with ReLU) is selected to further investigate the latent parameters (i.e., $\mu$, $\mu \oplus \sigma$, or the reparameterized $\mu$ and $\sigma$), as presented in the bottom row. The results are reported for three dense layers, which yield the highest performance. While FactorVAE is trained only once, the search process for the compressed classifier is repeated 10 times, and standard deviation results are provided. FactorVAE failed to converge in only one case ($d=32$, indoor large-scale dataset). As the dimensionality of $z$ decreases, the final accuracy declines only marginally. For the reparameterization approach, even a small latent dimension ($z = 16$) proves to be effective. Moreover, concatenating $\mu \oplus \sigma$ achieves higher performance compared to using only $\mu$.

\begin{figure}[!t]
    \centering
	\begin{minipage}[t]{0.325\linewidth}
        \centering
        \includegraphics[trim=10 10 10 10, clip, width=1.0\linewidth]{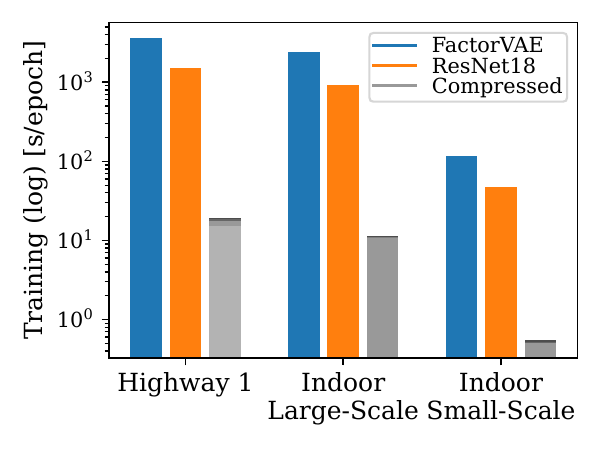}
    \end{minipage}
    \hfill
	\begin{minipage}[t]{0.325\linewidth}
        \centering
        \includegraphics[trim=10 10 10 6, clip, width=1.0\linewidth]{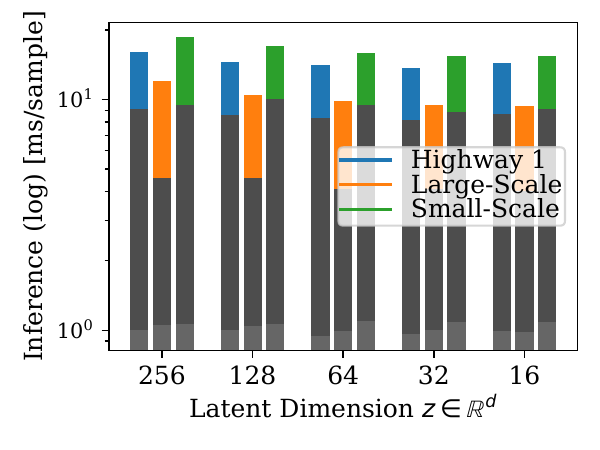}
    \end{minipage}
    \hfill
	\begin{minipage}[t]{0.325\linewidth}
        \centering
        \includegraphics[trim=10 10 10 10, clip, width=1.0\linewidth]{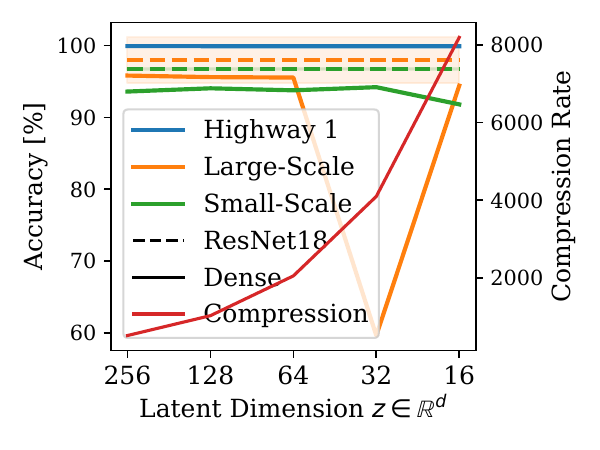}
    \end{minipage}
    \caption{Comparison of training time, inference time, accuracy, and compression rate for ResNet18 and the compressed model.}
    \label{figure_times}
\end{figure}

\textbf{Inference Time \& Compression Rate.} Figure~\ref{figure_times} provides an overview of training and inference times, along with a comparison of accuracy and compression rates across three datasets. While FactorVAE exhibits a slightly longer training duration than ResNet18, the compressed model achieves substantially faster training. However, the primary focus is on minimizing inference times to enable low-latency communication. Due to compression, the dense classifier achieves significantly shorter inference times than FactorVAE, irrespective of the latent dimension. With a compression rate ranging from 512 to 8,192, the compressed \textit{dense} model maintains high classification accuracy, incurring only a 0.04\%p, 2.3\%p, and 2.7\%p reduction compared to \textit{ResNet18}.

\section{Conclusion}
\label{label_conclusion}

We propose the use of factorized and generative VAEs for disentanglement and data augmentation in the compressed sensing of GNSS interference spectrograms. Our approach effectively extracts relevant latent features and enables interpolation across signal power levels, bandwidths, and distances, thereby enhancing model generalization through data augmentation. Experimental evaluations on two indoor and two real-world datasets demonstrate a compression rate of up to 8,192, achieving an accuracy range of 99.92\%, 95.63\%, and 94.05\% comparable to ResNet18 (99.96\%, 97.93\%, 96.75\%). This facilitates more efficient communication within distributed systems by enabling the deployment of smaller, faster models and the transmission of compressed information. This enables the deployment of models directly on GNSS receivers with high data acquisition rates. Future research will involve evaluating these models on receiver hardware to support situational awareness in open environments, thereby facilitating the development of crowdsourced and federated learning systems.

\section*{Acknowledgments}
\small This work has been carried out within the DARCII project, funding code 50NA2401, sponsored by the German Federal Ministry for Economic Affairs and Climate Action (BMWK) and supported by the German Aerospace Center (DLR), the Bundesnetzagentur (BNetzA), and the Federal Agency for Cartography and Geodesy (BKG).

\bibliography{ICL2025}
\bibliographystyle{IEEEtran}

\end{document}